\pdfoutput=1

\documentclass[11pt]{article}

\usepackage[preprint]{acl}

\usepackage{times}
\usepackage{latexsym}

\usepackage[T1]{fontenc}

\usepackage[utf8]{inputenc}

\usepackage{microtype}

\usepackage{inconsolata}

\usepackage{graphicx}

\usepackage{algorithm}
\usepackage{algpseudocode}
\usepackage{ragged2e}
\usepackage{amsmath}
\usepackage{amssymb}

\usepackage{setspace}
\usepackage{graphicx}
\usepackage{subcaption}
\usepackage{booktabs}

\usepackage{xspace}
\usepackage{todonotes}
\newcommand{\algo}{\textsc{DeFT-X}\xspace}

\title{\algo: Denoised Sparse Fine-Tuning for \\Zero-Shot Cross-Lingual Transfer}

\author{Sona Elza Simon \\
    Indian Institute of Technology Bombay\\
    Mumbai, India \\
    \texttt{sona.simon@iitb.ac.in} \\\And
    Preethi Jyothi \\
    Indian Institute of Technology Bombay\\
    Mumbai, India \\
    \texttt{pjyothi@cse.iitb.ac.in} \\}

\begin{document}
\maketitle
\begin{abstract}
Effective cross-lingual transfer remains a critical challenge in scaling the benefits of large language models from high-resource to low-resource languages. Towards this goal, prior studies have explored many approaches to combine \emph{task knowledge} from task-specific data in a (high-resource) source language and \emph{language knowledge} from unlabeled text in a (low-resource) target language. One notable approach proposed composable sparse fine-tuning (SFT) for cross-lingual transfer that learns task-specific and language-specific sparse masks to select a subset of the pretrained model's parameters that are further fine-tuned. These sparse fine-tuned vectors (SFTs) are subsequently composed with the pretrained model to facilitate zero-shot cross-lingual transfer to a task in a target language,  using only task-specific data from a source language. These sparse masks for SFTs were identified using a simple magnitude-based pruning. In our work, we introduce \algo, a novel composable SFT approach that denoises the weight matrices of a pretrained model before magnitude pruning using singular value decomposition, thus yielding more robust SFTs. We evaluate \algo on a diverse set of extremely low-resource languages for sentiment classification (NusaX) and natural language inference (AmericasNLI) and demonstrate that it performs at par or outperforms SFT and other prominent cross-lingual transfer baselines.
\end{abstract}

\section{Introduction}

Pretrained language models (LMs) are the de-facto choice for NLP, achieving state-of-the-art results across diverse benchmarks. However, effectively adapting these models to specific tasks remains a challenge owing to their large model sizes and the substantial training costs incurred during full fine-tuning. Furthermore, full fine-tuning approaches are prone to issues such as catastrophic forgetting and negative interference when adapted to multiple tasks. To mitigate these challenges, parameter-efficient fine-tuning (PEFT) techniques are a popular choice~\cite{pfeiffer2024modulardeeplearning}. These approaches include sparse fine-tuning (SFT) that refers to identifying a sparse subnetwork of the full model to train, and adapter-based methods that insert additional trainable modules while keeping the original model parameters fixed \cite{pfeiffer-etal-2020-mad, hu2022lora}.

Multilingual NLP introduces an additional layer of complexity, especially in the context of low-resource languages. A key objective of recent work in multilingual NLP has been to facilitate cross-lingual transfer by leveraging high-resource language data to improve performance on low-resource languages. Zero-shot cross-lingual transfer refers to the more constrained setting of having access only to task-specific labeled data in a high-resource (source) language and no labeled data in a low-resource (target) language. PEFT-based approaches such as MAD-X \citep{pfeiffer-etal-2020-mad} and LT-SFT \citep{ansell-etal-2022-composable} are designed to support zero-shot cross-lingual transfer. In MAD-X, a task adapter is learned using labeled data in a source language and a language adapter is learned using unlabeled data in a target language. To achieve zero-shot cross-lingual transfer, the task adapter in the source language is combined with the language adapter in the target language. LT-SFT (Lottery Ticket Sparse Fine-Tuning) removes the need for new trainable modules as in adapters by first identifying sparse subnetworks (a.k.a. ``lottery tickets") within the model using magnitude pruning. Fine-tuning these subnetworks while keeping the rest of the model frozen yields sparse vectors; these vectors can be estimated independently to learn task-specific and language-specific knowledge. These vectors are composed via simple addition to obtain the final model for zero-shot cross-lingual transfer.

\begin{figure*}
    \centering
    \includegraphics[width=0.98\textwidth]{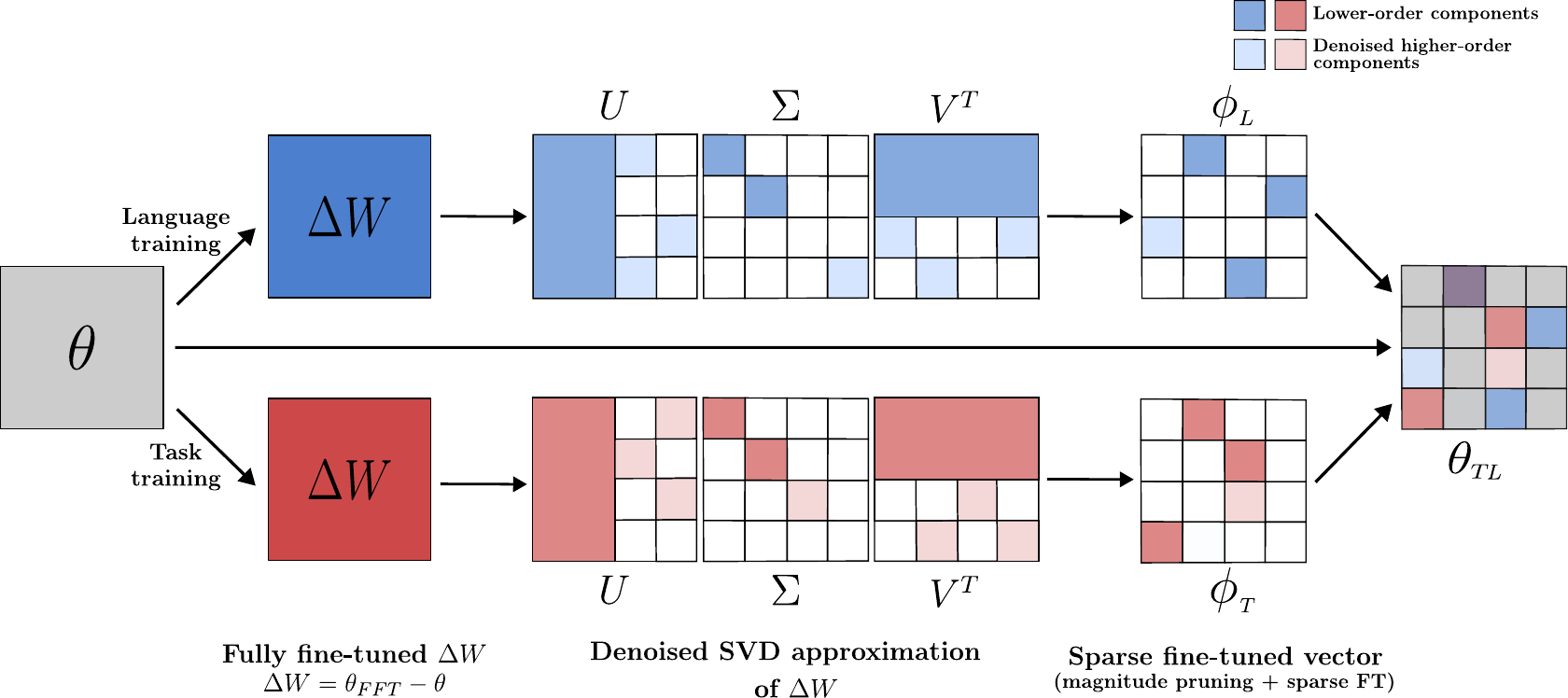}
    \caption{A graphical representation of \algo. The pretrained model $\theta$ (gray, left) undergoes full fine-tuning to obtain $\theta_{\text{FFT}}$. The difference $\Delta W$ (blue and red, left) captures the magnitude difference between $\theta$ and $\theta_{\text{FFT}}$. Each weight matrix in $\Delta W$ is denoised by pruning higher-order components (i.e., lower singular value components) while retaining lower-order components (i.e., high singular value components). The denoised $\Delta W$ is then magnitude-pruned and sparsely fine-tuned to produce $\phi$. Finally, the language-specific component $\phi_L$ and task-specific component $\phi_T$ are combined via addition to form the target language-task model $\theta_{\text{TL}}$(left).}
    \label{fig:algo}
\end{figure*}

In this work, we propose \algo, a novel sparse composable approach for zero-shot cross-lingual transfer. Adopting the LT-SFT template, \algo aims to identify subnetworks for further finetuning using a low-rank approximation and improves the quality of sparse fine-tuned vectors by denoising higher-order components. Specifically, we use Singular Value Decomposition (SVD) \citep{sharma2023truththereimprovingreasoning, zhao2025lowranksparsemodelmerging} to decompose each weight matrix into lower and higher-order components (corresponding to high and low singular values, respectively). The higher-order components are denoised and added to the SVD lower-order components, resulting in a signal-amplified weight matrix (refer to $\S$\ref{sec:methodology}). Next, we apply magnitude pruning to identify a more effective sparse subnetwork by removing noise, followed by fine-tuning to obtain the final sparse fine-tuned vectors (detailed in $\S$\ref{sec:methodology}). These vectors are finally composed via simple addition for effective cross-lingual transfer (as shown in $\S$\ref{sec:xlinguallearning}). An illustration of \algo is shown in Figure \ref{fig:algo}.

We evaluate the zero-shot cross-lingual performance of \algo on sentiment analysis (SA) and natural language inference (NLI) tasks for extremely low-resource Indonesian/indigenous languages in the NusaX/AmericasNLI benchmarks, respectively. Our results demonstrate that \algo outperforms state-of-the-art baselines. Our findings highlight the importance of each step in the \algo pipeline -- denoising the model before selecting a sparse subnetwork, enforcing sparsity after denoising and fine-tuning the subnetwork, particularly in low-resource settings.

\section{Background}
\subsection{Lottery Ticket Hypothesis (LTH)}
The \emph{Lottery Ticket Hypothesis (LTH)} \citep{frankle2018the, malach2020proving} states that within a randomly initialized, dense neural network, there exists a subnetwork referred to as a \emph{winning ticket} that when trained in isolation with its original initialization, can achieve comparable or even superior performance to the full network. 
Simple pruning techniques such as magnitude-based pruning can be used to identify such trainable subnetworks in fully connected and convolutional networks \cite{10.5555/2969239.2969366, DBLP:journals/corr/HanMD15, 8100126}.

To identify a winning ticket, a neural network \( f(x; \theta^{(0)}) \) is first initialized with random parameters \( \theta^{(0)} \). The network is trained for \( j \) iterations, resulting in parameters \( \theta^{(j)} \). Next, \( p\)\% of the smallest-magnitude weights are pruned, creating a binary mask \( m \). Finally, the remaining parameters are reset to their original values from \( \theta^{(0)} \) to yield the winning ticket \( f(x; m \odot \theta^{(0)}) \). However, if a winning ticket’s parameters are randomly re-initialized, its performance deteriorates, highlighting the importance of proper initialization for effective training. This process of pruning and resetting helps uncover subnetworks that retain strong learning abilities when trained in isolation. 

\subsection{Lottery Ticket Sparse Fine-Tuning (LT-SFT)}
Inspired by the LTH, \citet{ansell-etal-2022-composable} proposed a lottery ticket-based algorithm (LT-SFT) for efficient zero-shot cross-lingual transfer learning for low-resource languages. LT-SFT proposes a parameter-efficient fine-tuning approach that is both modular like adapters \citep{pfeiffer2021adapterfusionnondestructivetaskcomposition, pfeiffer-etal-2020-mad} (i.e., it can be easily combined to adapt a model to different knowledge sources) and expressive like sparse fine-tuning (i.e., it changes the behavior of all components). LT-SFT outperformed the state-of-the-art MAD-X adapter-based approach for low-resource cross-lingual transfer; the algorithm is detailed below.

\paragraph{LT-SFT: Generate Sparse Fine-tuned Vectors (SFTs).}
The LT-SFT algorithm generates sparse fine-tuned vectors (SFTs) in two phases: (1) Pretrained model parameters $\theta^{(0)}$ are fine-tuned on a target task or language to obtain $\theta^{(1)}$. From the top-$K$ winning tickets based on the greatest absolute difference $|\theta^{(1)}_i - \theta^{(0)}_i|$, a binary mask $b \in \{0,1\}$ is constructed where $b = 1$ for selected parameters and $b = 0$ otherwise. (2) All the parameters are reset to  $\theta^{(0)}$ and the model is retrained with keeping all non-winning parameters frozen using $b$. The resulting fine-tuned parameters $\theta^{(2)}$ yield the sparse fine-tuned vector (SFT) $\phi = \theta^{(2)} - \theta^{(0)}$.

\paragraph{LT-SFT: Zero-shot Transfer using SFTs.}
For each target language $l$, a language-specific SFT $\phi^{l}_{L}$ is estimated via the masked language modeling (MLM) objective on text from language $l$, initialized with the pretrained model weights $\theta_0$.
For each task $t$, a task-specific SFT $\phi^{t}_{T}$ is learned by training LT-SFT on annotated data in the source language $s$. For learning the task SFT, the LT-SFT algorithm first adapts the pre-trained model to the source language by adding the source language SFT $\phi^{s}_{L}$ to the model initialization i.e, $\theta_0 + \phi^{s}_{L}$. The model is then trained on the task to obtain updated parameters $\theta^{'}$. Finally, the task-specific SFT is computed by removing the source language SFT: $\phi^{t}_{T} = \theta^{'} - (\theta^{(0)} + \phi^{s}_{L})$. During task training, a classifier head is learned and fully fine-tuned in both phases of LT-SFT, with its random initialization reset at the start of each phase. For zero-shot cross-lingual transfer, the language-specific SFT $\phi_L$ and task-specific SFT $\phi_T$ are composed with the pretrained model as $\theta_{TL} = \theta^{(0)} +\phi_T +\phi_L$ to yield the target language-task model. The classifier head trained for the task is stacked on top to obtain the final model.

\section{Methodology}

\subsection{\algo: Motivation}
\label{sec:methodology}

Sparse fine-tuned vectors enable parameter-efficient cross-lingual transfer, but its effectiveness depends heavily on the quality of the identified sparse vectors. In LT-SFT, these vectors might capture noise or irrelevant information. \algo mitigates this by introducing a low-rank denoising step prior to sparse fine-tuning. \algo prunes the higher-order (lower singular value) components from the model weights, which are more likely to capture uninformative or noisy artefacts. The resulting denoised sparse vectors lead to more effective and robust transfer, especially in low-resource language scenarios.

\begin{table*}[t]
\setlength{\tabcolsep}{6pt}
\begin{tabular}{p{2.2cm}p{2cm}p{4.5cm}p{1.6cm}p{3.5cm}}
\toprule
\textbf{Task} & \textbf{Target Dataset} & \textbf{Target Languages} & \textbf{Source \newline Language} & \textbf{Source Task \newline Dataset} \\ \midrule
Sentiment \newline Analysis (SA) & NusaX \citep{winata-etal-2023-nusax} & Acehnese, Balinese, Banjarese, Madurese, Minangkabau & Indonesian & SMSA \citep{Purwarianti_2019, wilie-etal-2020-indonlu} \\ \midrule
Natural \newline Language \newline Inference (NLI) & AmericasNLI \citep{ebrahimi-etal-2022-americasnli} & Aymara, Asháninka, Bribri, Guarani, Náhuatl, Otomí, Quechua, Rarámuri, Shipibo-Konibo, Wixarika & English & MultiNLI \citep{williams-etal-2018-broad} \\ \bottomrule
\end{tabular}
\caption{Details of the tasks, datasets, and languages involved in our zero-shot cross-lingual transfer evaluation. All target languages are low-resource and were unseen during XLM-R pretraining. All the training data was obtained from the authors of \citet{ansell-etal-2022-composable, ansell-etal-2023-unifying}. Further details are provided in Appendix \ref{app:dataset}.}
\label{table:dataset_details}
\end{table*}

\paragraph{Denoising using Low-Rank Approximation.} 
In \algo, we start with identifying the `winning tickets' for efficient cross-lingual transfer. We compute the difference between the pretrained model parameters $\theta^{(0)}$ and the fully fine-tuned model parameters $\theta^{(1)}$ to obtain  
\begin{equation}
    \delta = \theta^{(1)} - \theta^{(0)}
\end{equation}
To extract the winning tickets in $\delta$, we first obtain a low-rank approximation \citep{zhao2025lowranksparsemodelmerging} by decomposing each weight matrix $W \in  \mathbb{R}^{m \times n}$ in $\delta$ using Singular Value Decomposition (SVD):  
\begin{equation}
    W = U\Sigma V^{T}
\end{equation}
where $U \in \mathbb{R}^{m \times m}$ and $V \in \mathbb{R}^{n \times n}$ are the left and right singular vector matrices of $W$ and $\Sigma \in \mathbb{R}^{m \times n}$ is the diagonal matrix of singular values. We construct a low-rank approximation of $W$ by retaining its lower-order (high singular value) components while pruning noise from the higher-order (low singular value) components:  
\begin{equation}
    \begin{aligned}
        L &= U_r\Sigma_r V_r^{T}, &\quad S &= m \odot (W - L)
    \end{aligned}
\end{equation}
where $U_r$ and $V_r$ denote the first $r$ columns of $U$ and $V$, respectively. The matrix $L$ captures the lower-order components, while $S$ represents the denoised higher-order components. The mask $m$ is defined as:
\begin{equation}
    m_i =
    \begin{cases} 
        1,& \text{if } i \in \text{Top-}n\text{-indices of } |W - L| \\
        0,& \text{otherwise}.
    \end{cases}
\end{equation}

Here, the higher-order components $(W - L)$ are pruned using magnitude-based selection, where the mask $m$ retains the $n$ largest absolute values while discarding the rest. The final low-rank approximation of $W$ is reconstructed as:  
\begin{equation}
    \label{eq:lowrank}
    W \approx L + S
\end{equation}

\paragraph{Sparse Fine-Tuning.}
After obtaining the low-rank approximation of $\delta$ by altering each matrix as shown in Eqn~\ref{eq:lowrank}, we apply magnitude pruning to obtain a sparse structure for efficient model composition. Sparsity is crucial to avoid destructive interference during model composition. For magnitude pruning, we construct a binary mask $\mu \in \{0,1\}$ where $\mu = 1$ for the top-k entries in the denoised $\delta$ with the highest absolute values, and $\mu = 0$ otherwise. We reset all the parameters to pretrained weights $\theta^{(0)}$ and perform sparse fine-tuning while keeping the non-winning parameters frozen using $\mu$. The resulting fine-tuned parameters $\theta^{(2)}$ yield the sparse fine-tuned vector $\phi = \theta^{(2)} - \theta^{(0)}$.

\subsection{Zero-shot Cross-lingual Transfer Learning}
\label{sec:xlinguallearning}

The sparse fine-tuned vector obtained from \algo can be easily composed similar to LT-SFT. For each target language $l$, a language-specific vector $\phi^{l}_{L}$ is learned via the \algo algorithm using masked language modeling (MLM) with unlabeled text from language $l$.
For each task $t$, a task-specific vector $\phi^{t}_{T}$ is learned by training \algo on annotated data in the source language $s$. Similar to LT-SFT, for task vector training we first adapt the model to the source language by applying the language SFT of the source language $\phi^{s}_{L}$ and then train the model on the task. During task training, a classifier head is also learned, which is randomly initialized and fully fine-tuned. For zero-shot cross-lingual transfer, these language and task-specific vectors can be easily composed with the pretrained model to obtain the desired language-task model as follows: $\theta_{TL} = \theta^{(0)} +\phi_T +\phi_L$, where $\theta_{TL}$ represents the target language-task model. The classifier head learned for task $t$ is stacked on top to obtain the final model that is used for zero-shot inference. 

The algorithm for \algo, including the steps for cross-lingual transfer, is given in Appendix  \ref{app:algorithm}.

\section{Experimental Setup}
We evaluate zero-shot cross-lingual transfer learning on two low-resource benchmarks: Sentiment Analysis (NusaX) and Natural Language Inference (AmericasNLI). Table \ref{table:dataset_details} summarizes the experimental setup, including the datasets and target languages considered.

\subsection{Baselines}
Our primary baseline is LT-SFT \citep{ansell-etal-2022-composable}, the current state-of-the-art framework for zero-shot cross-lingual transfer. We train all task-specific and target-language SFTs using the dataset provided by the LT-SFT authors and report the corresponding results. Additionally, we compare against the MAD-X 2.0 variant \citep{pfeiffer-etal-2021-unks}, using previously reported MAD-X results from~\citet{ansell-etal-2022-composable}.

\begin{table*}[t]
    \renewcommand{\arraystretch}{1.1}
    \centering
    \begin{tabular}{c|c|ccccc|c}
    \toprule
        \textbf{Model} & \textbf{Method} & \textbf{mad} & \textbf{bjn} & \textbf{ban} & \textbf{ace} & \textbf{min} & \textbf{Avg.}\\ \midrule
        XLM-R$_{\text{BASE}}$ & MAD-X \citep{ansell-etal-2022-composable} & 68.5 & 77.6 & 78.0 & 74.9 & 79.9 & 75.8 \\ 
        & LT-SFT & 79.0 & 82.7 & 80.4 & 75.7 & 83.0 & 80.2 \\ 
        & \algo ($r_l$ = $r_t$ = 90\% var) & \underline{\textbf{80.5}} & \textbf{83.5} & \underline{\textbf{82.7}} & 74.2 & \underline{\textbf{85.2}} & \textbf{81.2} \\
        & \algo ($r_l$ = $r_t$ = 100) & \textbf{79.8} & \textbf{83.8} & \textbf{81.4} & \underline{\textbf{76.8}} & \textbf{85.1} & \underline{\textbf{81.4}} \\
        & \algo ($r_l$ = 90\% var; $r
        _t$= 100)& \textbf{79.8} & \underline{\textbf{84.1}} & \textbf{82.2} & \textbf{76.3} & \textbf{83.8} & \textbf{81.2} \\
         & \algo ($r_l$ = 100; $r
        _t$= 90\% var)& \textbf{79.3} &\textbf{82.8} & \textbf{81.5} & 75.1 & \textbf{85.0} & \textbf{80.7} \\
        \midrule
        XLM-R$_{\text{LARGE}}$ & LT-SFT & 74.9 & 86.7 & 83.4 & \underline{80.0} & 87.1 & 82.4 \\ 
        & \algo ($r_l$ = $r_t$ = 90\% var)& 74.0 & 86.0 & 82.4 & 78.9 & \underline{\textbf{89.0}} & 82.1 \\
        & \algo ($r_l$ = $r_t$ = 200) & \textbf{76.1} & \underline{\textbf{87.2}} & \underline{\textbf{84.8}} & 79.2 & \textbf{88.7} & \underline{\textbf{83.2}} \\
        & \algo ($r_l$ = 90\% var; $r_t$= 200)& \textbf{75.8} & \textbf{87.0} & 82.7 & 77.9 & \textbf{87.8} & 82.2 \\
        & \algo ($r_l$ = 200; $r_t$= 90\% var)& \underline{\textbf{76.3}} & 86.0 & \textbf{84.6} & 79.9 & \textbf{88.2} & \textbf{83.0} \\  \bottomrule
    \end{tabular}
    \caption{Zero-shot cross-lingual transfer evaluation (F1-Score) on SA task (NusaX) using  XLM-R$_{\text{BASE}}$ and XLM-R$_{\text{LARGE}}$. XLM-R$_{\text{LARGE}}$ numbers are without $\phi_L^s$ initialization for task. For MAD-X baseline, we present the numbers reported in \citet{ansell-etal-2022-composable}. Here, $r_l$ and $r_t$ denote the rank used for language and task sparse vectors respectively. \textbf{Bold} indicates performance surpassing the baselines, while \underline{underline} denotes the best performance.}
    \label{tab:SA_results}
\end{table*}

\subsection{Training Setup}
For our experiments, we use the pretrained XLM-R$_{\text{BASE}}$ and XLM-R$_{\text{LARGE}}$ models and conduct all training on a single NVIDIA A100 80GB GPU. To ensure fair comparisons, we adopt the same training setup used in LT-SFT \citep{ansell-etal-2022-composable, ansell-etal-2023-unifying} for both language and task training. Implementation details including the training steps, optimizer settings, etc. are detailed in Appendix~\ref{app:implementation}.

For language SFTs trained using the MLM objective, the number of trainable parameters $k$ is set to $7.6M$ (i.e., $2.8\%$ and $1.4\%$ of the parameters in XLM-R$_{\text{BASE}}$ and XLM-R$_{\text{LARGE}}$, respectively). And for task SFTs, $k$ is set to $14.2M$ (i.e., $5.2\%$ and $2.6\%$ of the parameters in XLM-R$_{\text{BASE}}$ and XLM-R$_{\text{LARGE}}$ respectively). This choice of $k$ is consistent with the LT-SFT baseline, and the value of $k$ was selected such that it matches the number of parameters in the MAD-X adapters.%
\footnote{The LT-SFT \cite{ansell-etal-2022-composable} uses a reduction factor of $2$ and $1$ for language and task MAD-X adapters, respectively.} During task adaptation, we always apply the source language SFT from LT-SFT \citep{ansell-etal-2022-composable} to the XLM-R$_{\text{BASE}}$ model , while the XLM-R$_{\text{LARGE}}$ model is trained without source language initialization due to the unavailability of source language SFT for the large model.

\paragraph{Denoising using Low-Rank Approximation.}
We denoise all trainable weight matrices of the model, except for the bias terms; we apply direct magnitude-based pruning for the latter. The SVD operations on each weight matrix are computed in parallel. We explore two methods for selecting the appropriate rank to separate higher- and lower-order components in the matrix. Following \citet{chang-etal-2022-geometry}, we choose the rank $r$ that captures 90\% of the total variance for each matrix i.e, 
 we first calculate the total variance from all singular values in the matrix, then retains the minimum number of singular vectors needed such that their cumulative variance reaches 90\% of the total variance. Additionally, we investigate a uniform rank selection across layers by setting a uniform rank $r$ = \{100, 200, 300\}, based on our empirical observation that most rank values derived from the 90\% variance criterion fall within this range. To denoise the higher-order components, we apply magnitude pruning while retaining only 5\% of the higher-order components in each matrix.

\begin{table*}[t]
    \renewcommand{\arraystretch}{1.1}
    \setlength{\tabcolsep}{4.25pt}
    \centering
    \begin{tabular}{@{}c|cccccccccc|c@{}}
    \toprule
        \textbf{Method} & \textbf{bzd} & \textbf{oto} & \textbf{hch} & \textbf{tar} & \textbf{cni} & \textbf{shp} & \textbf{aym} & \textbf{gn} & \textbf{nah} & \textbf{quy} & \textbf{Avg.}\\ \midrule
        \multicolumn{12}{c}{\textbf{XLM-R$_{\text{BASE}}$}} \\ \midrule
        MAD-X \citep{ansell-etal-2022-composable}& 44.0 & 46.8 & 41.5 & 43.9 & 47.6 & 48.9 & 58.8 & 63.5 & 53.7 & 58.3 & 49.5 \\
        LT-SFT & 43.6 & 45.6 & 42.9 & \underline{44.8} & 47.5 & 49.2 & \underline{60.4} & 63.3 & 50.9 & 62.1 & 51.0 \\ 
        \algo ($r_l$ = $r_t$ = 90\% var) & 43.3 & \underline{\textbf{47.6}} & \textbf{44.0} & 41.1 & 45.6 & 49.0 & 58.8 & 63.0 & 49.6 & 61.9 & 50.4 \\ 
       \algo ($r_l$ = $r_t$ = 200) & 42.9 & 43.8 & \underline{\textbf{45.6}} & 44.1 & \textbf{48.0} & \textbf{49.3} & 58.5 & \textbf{63.6} & \underline{\textbf{53.9}} & 61.7 &\textbf{51.2} \\
        \algo ($r_l$ = 90\% var; $r_t$= 200) & 42.4 & 45.4 & \textbf{44.0} & 42.8 & 46.5 & 48.8 & 58.1 & 63.5 & 51.5 & \underline{\textbf{62.8}} & 50.6 \\
        \algo ($r_l$ = 200; $r_t$= 90\% var)& \underline{\textbf{44.3}} & 44.5 & \textbf{43.6} & 44.0 & \underline{\textbf{48.1}} & \underline{\textbf{50.5}} & 58.1 & \underline{\textbf{64.5}} & 53.2 & \textbf{62.3} & \underline{\textbf{51.3}} \\ \midrule
        \multicolumn{12}{c}{\textbf{XLM-R$_{\text{LARGE}}$}} \\ \midrule
        LT-SFT & 43.9 & \underline{42.1} & 45.2 & 42.5 & 46.9 & 48.7 & 58.0 & 54.7 & 39.6 & 50.1 & 47.2 \\ 
        \algo ($r_l$ = $r_t$ = 90\% var) & \textbf{44.5} & 40.5 & 44.9 & \underline{\textbf{43.5}} & 46.8 & \textbf{50.8} & 57.5 & 52.5 & 37.8 & \textbf{50.8} & 47.0 \\ 
        \algo ($r_l$ = $r_t$ = 300) & \textbf{44.5} & 41.4 & 45.2 & \textbf{43.1} & 46.1 & \textbf{49.7} & \underline{\textbf{58.8}} & \underline{\textbf{55.9}} & 39.3 & \textbf{51.2} & \textbf{47.5} \\ 
        \algo ($r_l$ = 90\% var; $r_t$= 300) & \textbf{44.5} & 41.2 & 43.9 & \underline{\textbf{43.5}} & 46.7 & \underline{\textbf{51.3}} & 57.5 & \textbf{55.2} & \underline{\textbf{40.4}} & \underline{\textbf{51.6}} & \underline{\textbf{47.6}} \\ 
        \algo ($r_l$ = 300; $r_t$= 90\% var) & \underline{\textbf{44.8}} & 40.1 & \underline{\textbf{45.6}} & \underline{\textbf{43.5}} & \underline{\textbf{47.2}} & \textbf{50.7} & \textbf{58.5} & \underline{\textbf{55.9}} & 38.9 & 49.7 & \textbf{47.5} \\
        \bottomrule
    \end{tabular}
    \caption{Zero-shot cross-lingual transfer evaluation (accuracy) on NLI task (AmericasNLI) using  XLM-R$_{\text{BASE}}$ and XLM-R$_{\text{LARGE}}$. XLM-R$_{\text{LARGE}}$ numbers are without $\phi_L^s$ initialization for task. For MAD-X baseline, we present the numbers reported in \citet{ansell-etal-2022-composable}. Here, $r_l$ and $r_t$ denote the rank used for language and task sparse vectors respectively. \textbf{Bold} indicates performance surpassing the baselines, while \underline{underline} denotes the best performance.}
    \label{tab:NLI_results}
\end{table*}

\subsection{Evaluation Datasets}
As shown in Table~\ref{table:dataset_details}, NusaX is an SA benchmark covering five low-resource Indonesian languages.%
\footnote{There are four other test languages in NusaX. However, no monolingual corpora were available for these languages; hence, these are omitted from our results and LT-SFT.}
There are 400 test samples in each language to be classified as either positive, negative or neutral. AmericasNLI \citep{ebrahimi-etal-2022-americasnli} is an extension of XNLI \citep{conneau-etal-2018-xnli} to 10 low-resource indigenous languages of the Americas with 750 test samples each. We use F1-score and accuracy as the evaluation metrics for SA and NLI, respectively. 

\section{Results and Discussion}
We report F1 scores for NusaX using both methods of rank selection: 90\% variance and uniform rank $r$, as shown in Table \ref{tab:SA_results}. For XLM-R$_{\text{BASE}}$, we use uniform $r$ = 100 and for XLM-R$_{\text{LARGE}}$ we use a higher rank $r$ = 200; these rank values were selected by observing the overall rank that covers 90\% variance. For XLM-R$_{\text{BASE}}$, \algo consistently outperforms the baselines MAD-X and LT-SFT. Our best-performing configuration, with $r_l$ = $r_t$ = 100, surpasses MAD-X and LT-SFT with average gains of 5.6 and 1.2, respectively. For XLM-R$_{\text{LARGE}}$, we report the numbers without source language initailization for the task since the source language sparse vectors are not available for XLM-R$_{\text{LARGE}}$ model. \algo also outperforms the baseline LT-SFT on various settings using XLM-R$_{\text{LARGE}}$.

\begin{table}[t]
    \centering
    \setlength{\tabcolsep}{2.5pt}
    \begin{tabular}{l|c|c}
    \toprule
        \textbf{Method} & \textbf{SA} & \textbf{NLI} \\ 
         & (F1-Score) & (Accuracy) \\ 
    \midrule
        LT-SFT  & 80.2 & 51.0\\ 
        LT-SFT w/o $\phi_L^s$ for task & 79.6 & 50.4\\ \midrule
        \algo & \textbf{81.4} & \textbf{51.3}\\ 
        \algo w/o  $\phi_L^s$ for task & \textbf{81.1} & \textbf{51.3}\\ 
    \bottomrule
    \end{tabular}
    \caption{Comparison of LT-SFT and \algo without source language initialization for task vectors, using XLM-R\textsubscript{BASE} on the SA and NLI tasks. We use $r_l$ = $r_t$ = 100 for SA and $r_l$ = 200, $r_t$= 90\% variance for NLI. \textbf{Bold} indicates best performing model.}
    \label{tab:source_initilization}
\end{table}

Similarly, we report the accuracy for AmericasNLI using both methods of rank selection: 90\% variance and uniform rank $r$ = 200 and $r$ = 300 for the base and large model respectively, as shown in Table \ref{tab:NLI_results}. 
XLM-R$_{\text{BASE}}$, with $r_l$ = 200 and $r_t$ = 90\% variance, surpasses MAD-X and LT-SFT by 1.8\% and 0.3\%, respectively, on average. We observed that the languages in AmericasNLI are more low-resource than those in NusaX and require a higher rank $r_l$ = 200 to capture useful lower-order components. However, both baseline methods and \algo show degraded performance with XLM-R$_{\text{LARGE}}$ compared to the base model. This could be attributed to stronger biases towards high-resource languages during XLM-R$_{\text{LARGE}}$'s pretraining that make it less amenable to adapt to extremely low-resource languages in AmericasNLI. Even in such challenging settings, \algo maintains a consistent albeit modest improvement over LT-SFT.

\begin{table*}[t]
    \centering
    \begin{tabular}{l|ccccc|c}
    \toprule
        \textbf{ Method} & \textbf{mad} & \textbf{bjn} & \textbf{ban} & \textbf{ace} & \textbf{min} & \textbf{Avg.} \\ 
    \midrule
        \algo & \underline{79.8} & \underline{83.8} & 81.4 & \underline{76.8} & \underline{85.1} & \underline{81.4} \\ 
        \quad w/o higher-order components & \underline{79.8} & 82.8 & \underline{82.1} & 73.4 & 84.9 & 80.6 \\ 
        \quad w/o magnitude pruning + sparse fine-tuning & 77.4 & 81.7 & 79.7 & 74.6 & 82.7 & 79.2 \\ 
        \quad w/o sparse fine-tuning & 72.0 & 75.4 & 67.7 & 57.4 & 81.4 & 70.8\\ 
    \bottomrule
    \end{tabular}
    \caption{Analyzing the impact of higher-order components, sparsity(magnitude pruning), and the necessity of re-training (sparse fine-tuning) in \algo using the NusaX dataset on XLM-R\textsubscript{BASE} with $r_l$ = $r_t$ = 100. \underline{Underline} denotes the best performance.}
    \label{tab:SA_ablation}
\end{table*}

In summary, we observe that denoising higher-order components before selecting the sparse subnetwork (via magnitude pruning) improves the quality of sparse vectors for composition. Both uniform rank selection and 90\% variance-based rank selection perform comparably well. However, very low-resource languages may benefit from a higher rank to capture more meaningful knowledge.

\paragraph{Source Language Initialization for Task Vectors.}
We analyze the impact of source language initialization on training task sparse vectors in Table \ref{tab:source_initilization}. Our findings indicate that the baseline LT-SFT is sensitive to source language initialization, as its performance drops in its absence. In contrast, our approach, \algo, maintains comparable performance even without source language initialization. This suggests that denoising higher-order components before selecting the subnetwork (via magnitude pruning) leads to a more robust network compared to LT-SFT, which relies solely on magnitude pruning for subnetwork selection.

\begin{figure}[t]
    \centering
    \includegraphics[width=\linewidth]{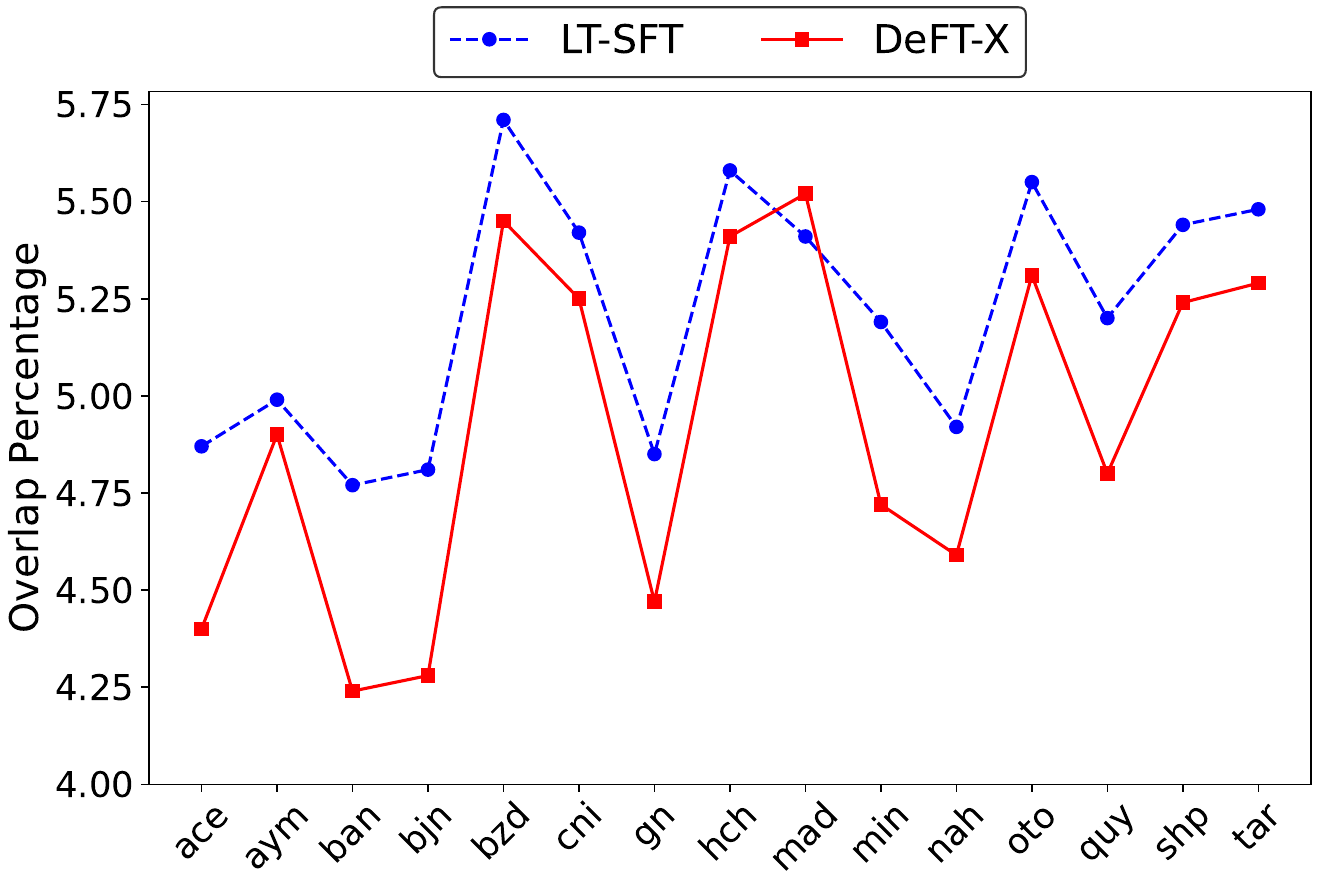}
    \caption{Comparing the overlap between the sparse language vectors and its corresponding task vectors. For \algo, we compare using $r_l$ = $r_t$ =100.}
    \label{fig:overlap_task_lang}
\end{figure}

\begin{figure}[t]
    \centering
    \includegraphics[width=\linewidth]{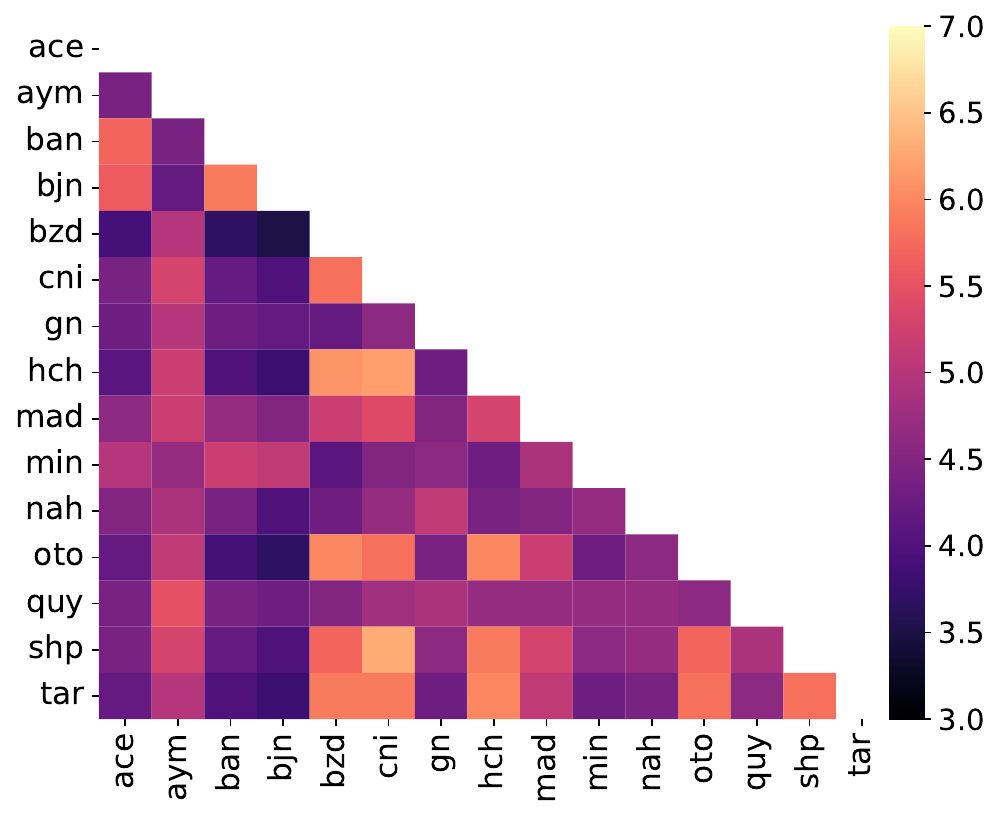}
    \caption{Overlap (in percentage) between the sparse language vectors of DeFT-X at $r_l$=100.}
    \label{fig:overlap_langs}
\end{figure}

\paragraph{Benefits of Denoising, Sparsity and Re-Training.} 
We investigate the need to retain de-noised higher-order parameters by comparing \algo with an alternative that entirely removes these parameters, as shown in Table \ref{tab:SA_ablation}. We observe that higher-order components contain useful information, making it essential to retain them after denoising. 
We also analyze the impact of magnitude pruning and sparse fine-tuning after denoising in Table \ref{tab:SA_ablation}. We find that magnitude pruning, when combined with sparse fine-tuning, effectively refines the parameter selection beyond denoising, leading to performance gains. In contrast, applying magnitude pruning alone without sparse fine-tuning results in a substantial drop in performance, underscoring the necessity of fine-tuning after pruning. This ablation study highlights the importance of each step in the \algo algorithm.

\paragraph{Parameter Overlap in Language and Task Vectors.} One of the key challenges in composing models for cross-lingual transfer is minimizing negative interference. Reducing parameter overlap can mitigate this interference by ensuring that language and task specific vectors learn distinct subnetworks. In Figure \ref{fig:overlap_task_lang}, we compare the parameter overlap between sparse fine-tuned language vectors and their corresponding task vectors for LT-SFT and \algo. Our findings show that \algo results in lower overlap across languages, indicating that denoising higher-order components helps remove redundancies while preserving language and task specific information. This leads to efficient cross-lingual transfer with minimized negative interference. We also analyze the overlap between language vectors in Figure~\ref{fig:overlap_langs} and find the overlap to be merely $\sim$5\%, suggesting that each language learns a distinct subnetwork within the pretrained model.

\section{Related Work}

\paragraph{Parameter-Efficient Fine-Tuning.}
Parameter-efficient fine-tuning (PEFT) adapts large pretrained models to downstream tasks with minimal trainable parameters. DiffPruning \citep{guo-etal-2021-parameter} learns sparse task-specific deltas using a differentiable $L_0$ penalty, while BitFit \citep{zaken2022bitfitsimpleparameterefficientfinetuning} restricts the updates to bias terms. Adapters \citep{pfeiffer2021adapterfusionnondestructivetaskcomposition} insert lightweight task-specific bottlenecks into Transformer layers, keeping the rest frozen. LoRA \citep{hu2022lora} introduces trainable low-rank decomposition matrices into each Transformer layer. Wanda \citep{sun2023simple} prunes weights based on the elementwise product of its magnitude and the corresponding input activation norm. These approaches significantly reduce the number of trainable parameters while preserving performance. Similarly, our approach \algo uses magnitude-based pruning, while further mitigating noise through low-rank denoising.

\paragraph{Task Arithmetic and Multi-Task Transfer Learning.} \citet{ilharco2023editingmodelstaskarithmetic} introduces task vectors obtained by subtracting the weights of a pretrained model from those of a fine-tuned model. These task vectors can be manipulated through arithmetic operations to steer model behavior. Approaches like \citet{ansell2024scalingsparsefinetuninglarge} and DARE \citep{10.5555/3692070.3694452} propose identifying better task vectors by dynamically dropping and learning parameters. Various approaches were proposed to mitigate the destructive interference during task arithmetic in multi-task setting like resolving sign conflicts while merging \citep{NEURIPS2023_1644c9af} and learning mutually sparse vectors \citep{lotto}. \algo can also be used in multi-task setting to obtain better task vectors by denoising the redundant task information and mitigating destructive interference.


\paragraph{Cross-Lingual Transfer Learning.}
Cross-lingual transfer learning improves task performance in low-resource languages by leveraging knowledge from high-resource languages. Approaches like MAD-X \citep{pfeiffer-etal-2020-mad} use modular language and task specific adapters for composable transfer. MAD-G \citep{ansell-etal-2021-mad-g} introduces a contextual parameter generator trained on typological features from URIEL to build efficient adapters for low-resource languages. LT-SFT \citep{ansell-etal-2022-composable} proposes sparse task and language vectors that can be arithmetically composed for zero-shot transfer. Subsequent works combine few-shot fine-tuning with LT-SFT \citep{ansell-etal-2023-unifying} and explore scaling in task arithmetic \citep{parovic-etal-2024-investigating}. \citet{ansell-etal-2023-distilling} proposes a bilingual distillation approach to extract language-specific models from massively multilingual transformers for cross-lingual transfer. Layer swapping for transferring linguistic knowledge to reasoning tasks has also been explored \citep{bandarkar2024layerswappingzeroshotcrosslingual}. Our approach \algo, builds on LT-SFT by improving sparse vector quality via SVD-based denoising.


\paragraph{Low-Rank Approximation using SVD.} Singular Value Decomposition (SVD) is widely used for low-rank approximation, aiding both efficiency and interpretability in neural networks. LASER \citep{sharma2023truththereimprovingreasoning} shows that pruning small singular components from Transformer weight matrices can improve reasoning by denoising internal representations without retraining. LoRS-Merging \citep{zhao2025lowranksparsemodelmerging} merges multilingual speech models using coarse-grained singular value pruning to retain essential structures and fine-grained magnitude pruning to remove redundancy. Our approach \algo, applies similar low-rank denoising to enhance the quality of sparse fine-tuned vectors.

\section{Conclusion and Future Work}
We introduced \algo, a composable, denoised, sparse fine-tuning approach for efficient zero-shot cross-lingual transfer. By leveraging SVD to denoise model weights, \algo identifies better subnetworks for sparse fine-tuning. We also explored different strategies for selecting the matrix rank during denoising. Compared to the state-of-the-art LT-SFT approach, \algo demonstrates improvements in SA and NLI tasks for low-resource languages. In future work, we plan to explore the broader applicability of \algo beyond cross-lingual transfer, including its potential in multimodal learning and domain adaptation.

\section*{Limitations}
While \algo shows promising results, our study has a few limitations. First, we evaluate the method only on encoder-only architectures, specifically transformer-based language models, leaving its effectiveness on decoder-only or encoder-decoder models unexplored. Second, our experiments are restricted to classification tasks (SA, NLI), and the applicability of \algo to other tasks remains to be investigated. Finally, the choice of the optimal rank for denoising via SVD is currently model and task-specific and requires manual tuning.


\bibliography{custom}

\clearpage
\appendix
\onecolumn  
\section{\algo Algorithm}
\label{app:algorithm}
\begin{algorithm*}
\caption{Cross-Lingual Transfer with \algo}
\begin{algorithmic}[1]
\setstretch{1.2} 
\Function {\algo}{$ D, L, \theta^{(0)}, \eta, k, n$}
    \State $\theta^{(1)} \leftarrow \theta^{(0)}$  \Comment{Full fine-tuning}
    \While{not converged} 
        \State $\theta^{(1)} \leftarrow \theta^{(1)} - \eta \nabla L(\theta^{(1)}, D)$
    \EndWhile 
    \vspace{0.25cm}
    \State $\delta \leftarrow \theta^{(1)} - \theta^{(0)}$ \Comment{SVD-based magnitude pruning}
    \For {each weight matrix $W \in \delta $} 
        \State $U\Sigma V^{T} \leftarrow SVD(W)$ 
        \State $L \leftarrow U_r\Sigma_r V^{T}_r $
        \State $m_i \leftarrow \begin{cases} 
          1 & \text{if } i \in \text{Top-}n\text{-indices of } |W - L| \\
          0 & \text{otherwise} \end{cases}$
        \State $S \leftarrow m \odot (W - L) $
        \State $W \leftarrow L + S$
    \EndFor
    \State $\mu_i \leftarrow \begin{cases} 
          1 & \text{if } i \in \text{Top-}k\text{-indices of } |\delta| \\
          0 & \text{otherwise} \end{cases}$
    \vspace{0.25cm}  
    \State $\theta^{(2)} \leftarrow \theta^{(0)}$ \Comment{Sparse fine-tuning}
    \While{not converged}  
        \State $\theta^{(2)} \leftarrow \theta^{(2)} - \mu \odot \eta \nabla L(\theta^{(2)}, D)$
    \EndWhile
    \State $\phi \leftarrow \theta^{(2)} - \theta^{(0)}$
    \State \Return $\phi$
\EndFunction
\vspace{0.5cm}
\Function {CrossLingual}{$D_{src}, D_{tar}, D_{task}, L_{task}, \theta^{(0)}, \eta, k, n$}
    \State $\phi_{src} \leftarrow \algo(D_{src}, L_{MLM}, \theta^{(0)}, \eta, k, n)$
    \State $\phi_{task} \leftarrow \algo(D_{task}, L_{task}, \theta^{(0)} + \phi_{src}, \eta, k, n)$
    \State $\phi_{tar} \leftarrow \algo(D_{tar}, L_{MLM}, \theta^{(0)}, \eta, k, n)$
    \State \Return $\theta^{(0)} + \phi_{task} + \phi_{tar}$
\EndFunction
\end{algorithmic}
\end{algorithm*}
\clearpage
\twocolumn

\section{Dataset Details}
\label{app:dataset}
We use publicly available data for training and evaluation with CC-BY-SA license. Table \ref{tab:dataset_details_app} provides a comprehensive overview of languages, their codes, linguistic families, and monolingual data sizes. For training the task vectors, we utilize the SMSA dataset \citep{Purwarianti_2019, wilie-etal-2020-indonlu} and the MultiNLI dataset \citep{williams-etal-2018-broad}.

Note: Since the NusaX dataset \citep{winata-etal-2023-nusax} was created through human translation of a subset of the SMSA dataset and we use the NusaX test set used by \citet{ansell-etal-2023-unifying} where they carefully removed every example from SMSA which appears in its original or modified form in the NusaX test set to avoid a data leak.

\section{Detailed Training Setup}
\label{app:implementation}
\paragraph{Language Adaptation.}The language vector is trained on the Masked Language Modeling (MLM) objective for the lesser of 100 epochs or 100,000 steps, using a batch size of 8 and a maximum sequence length of 256. However, training is subject to an absolute minimum of 30,000 steps, as 100 epochs appeared insufficient for some languages with very small corpora. Model checkpoints are evaluated every 1,000 steps on a held-out set comprising $5\%$ of the corpus, and the checkpoint with the lowest validation loss is selected at the end of training. We use the AdamW optimizer with an initial learning rate of $5\mathrm{e}{-5}$, which is linearly decayed to $0$ over the course of training. For language SFTs, the number of trainable parameters, $k$ is set to $7.6M$ (i.e., $2.8\%$ of the parameters in XLM-R$_{\text{BASE}}$). For adapters, the reduction factor (i.e., the ratio between model hidden size and adapter size) is set to 2 to ensure the number of trainable parameters matches that of SFT. Additionally, layer normalization parameters are kept fixed, while all other parameters remain trainable. For language adaptation, we apply L1 regularization with $\lambda = 0.1$. The specified training regime is applied consistently across both phases of LT-SFT.

\paragraph{Task Adaptation.}The task vector for SA is trained for 10 epochs with a batch size of 16, with checkpoint evaluation on the validation set every 250 steps, and the best checkpoint is taken at the end of training based on F1-score. The task vector for NLI is trained for 5 epochs with a batch size of 32, with checkpoint evaluation on the validation set every 625 steps, and the best checkpoint is taken at the end of training based on accuracy. Similarly to language adaptation, the task SFT training uses the AdamW optimizer with an initial learning rate of $5\mathrm{e}{-5}$, which is linearly decayed to $0$ over the course of training. A two-layer multi-class classification head is applied atop the XLM-R model output corresponding to the [CLS] token. For task SFTs, the number of trainable parameters $k$ is set to $14.2M$ (i.e., $5.1\%$ of the parameters in XLM-R$_{\text{BASE}}$). For adapters, the reduction factor is set to 1 to ensure the number of trainable parameters matches that of SFT. During task adaptation, we always apply the source language SFT from LT-SFT \citep{ansell-etal-2022-composable}.

\section{Ablation Experiments using Uniform Rank}
\label{app:uniorm_rank}
We conducted ablation experiments on the uniform rank variant by setting 
rank $r$ = \{100, 200\} for the SA and NLI tasks. The results are presented in Table \ref{tab:SA_uniform_rank} and Table \ref{tab:NLI_results_uniform_rank}.

\begin{table*}
    \renewcommand{\arraystretch}{1.1}
    \centering
    \begin{tabular}{c|c|c|c|c}
    \toprule
        \textbf{Task} & \textbf{Language} & \textbf{ISO Code} & \textbf{Family} & \textbf{Corpus size (MB)}\\ \midrule
        SA & Madurese & mad & Austronesian, Malayo-Sumbawan & 0.84 \\ 
         & Banjarese & bjn & Austronesian, Malayo-Sumbawan & 28.40 \\ 
         & Balinese & ban & Austronesian, Malayo-Sumbawan & 42.50 \\ 
         & Acehnese & ace & Austronesian, Malayo-Sumbawan & 89.70 \\ 
         & Minangkabau & min & Austronesian, Malayo-Sumbawan & 92.80 \\ \midrule 
        NLI & Bribri & bzd & Chibchan, Talamanca & 0.32 \\ 
         & Otomí & oto & Oto-Manguean, Otomian & 0.40 \\ 
         & Wixarika & hch & Uto-Aztecan, Corachol & 0.45 \\ 
         & Rarámuri & tar & Uto-Aztecan, Tarahumaran & 0.61 \\ 
         & Asháninka & cni & Arawakan & 1.40 \\ 
         & Shipibo-Konibo & shp & Panoan & 2.00 \\ 
         & Aymara & aym & Aymaran & 2.20 \\ 
         & Guarani & gn & Tupian, Tupi-Guarani & 6.60 \\ 
         & Náhuatl & nah & Uto-Aztecan, Aztecan & 7.70 \\ 
         & Quechua & quy & Quechuan & 16.00 \\ 
    \bottomrule
    \end{tabular}
    \caption{Details of languages, their codes, linguistic families, and monolingual data sizes used for MLM training of language vectors.}
    \label{tab:dataset_details_app}
\end{table*}

\begin{table*}
    \renewcommand{\arraystretch}{1.1}
    \centering
    \begin{tabular}{c|ccccc|c}
    \toprule
        \textbf{Method} & \textbf{mad} & \textbf{bjn} & \textbf{ban} & \textbf{ace} & \textbf{min} & \textbf{Avg.}\\ \midrule
        MAD-X \citep{ansell-etal-2022-composable}& 68.5 & 77.6 & 78.0 & 74.9 & 79.9 & 75.8 \\ 
        LT-SFT & 79.0 & 82.7 & 80.4 & 75.7 & 83.0 & 80.2 \\ \midrule
        \algo ($r_l$ = $r_t$ = 100) & \underline{\textbf{79.8}} & 
        \underline{\textbf{83.8}} & \textbf{81.4} & \underline{\textbf{76.8}} & \underline{\textbf{85.1}} & \underline{\textbf{81.4}} \\
        \algo ($r_l$ = $r_t$ = 200) & 78.4 & \textbf{83.0} & \textbf{81.0} & 73.9 & \textbf{83.4} & 80.0 \\
        \algo ($r_l$ = 100; $r_t$= 200)& \textbf{79.1} & \textbf{83.4} & \textbf{80.7} & 74.1 & \textbf{83.9} & 80.2 \\
        \algo ($r_l$ = 200; $r_t$= 100)& \textbf{78.5} & 82.5 & \underline{\textbf{83.0}} & \textbf{76.5} & \textbf{84.6} & \textbf{81.3} \\
        \bottomrule
    \end{tabular}
    \caption{Zero-shot cross-lingual transfer evaluation (F1-Score) on SA task (NusaX) for different uniform rank values using XLM-R$_{\text{BASE}}$. Here, $r_l$ and $r_t$ denote the rank used for language and task sparse vectors respectively. \textbf{Bold} indicates performance surpassing the baseline, while \underline{underline} denotes the best performance.}
    \label{tab:SA_uniform_rank}
\end{table*}

\begin{table*}
    \renewcommand{\arraystretch}{1.1}
    \setlength{\tabcolsep}{4.25pt}
    \centering
    \begin{tabular}{@{}c|cccccccccc|c@{}}
    \toprule
        \textbf{Method} & \textbf{bzd} & \textbf{oto} & \textbf{hch} & \textbf{tar} & \textbf{cni} & \textbf{shp} & \textbf{aym} & \textbf{gn} & \textbf{nah} & \textbf{quy} & \textbf{Avg.}\\ \midrule
        MAD-X \citep{ansell-etal-2022-composable}& 44.0 & \underline{46.8} & 41.5 & 43.9 & 47.6 & 48.9 & 58.8 & 63.5 & 53.7 & 58.3 & 49.5 \\
        LT-SFT & 43.6 & 45.6 & 42.9 & \underline{44.8} & 47.5 & 49.2 & \underline{60.4} & 63.3 & 50.9 & \underline{62.1} & 51.0 \\ \midrule
        
        \algo ($r_l$ = $r_t$ = 100) & 44.0 & 45.4 & 42.3 & 40.0 & 46.1 & \underline{\textbf{50.7}} & 60.3 & 63.0 & 52.0 & 61.1 & 50.5 \\ 
        
         \algo ($r_l$ = $r_t$ = 200) & 42.9 & 43.8 & \underline{\textbf{45.6}} & 44.1 & \underline{\textbf{48.0}} & \textbf{49.3} & 58.5 & \textbf{63.6} & \textbf{53.9} & 61.7 & \underline{\textbf{51.2}} \\ 
         
        \algo ($r_l$ = 100; $r_t$= 200) & 43.2 & 45.3 & \textbf{44.4} & 41.5 & 47.3 & \textbf{50.1} & \underline{60.4} & 62.0 & \underline{\textbf{54.6}} & 61.5 & 51.0 \\
        
        \algo ($r_l$ = 200; $r_t$= 100)& \underline{\textbf{44.7}} & 43.3 & 42.7 & 43.7 & 46.7 & \textbf{49.3} & 57.6 & \underline{\textbf{63.9}} & \textbf{53.9} & 61.6 & 50.7 \\
        \bottomrule
    \end{tabular}
    \caption{Zero-shot cross-lingual transfer evaluation (accuracy) on NLI task (AmericasNLI) for different uniform rank values using XLM-R$_{\text{BASE}}$. Here, $r_l$ and $r_t$ denote the rank used for language and task sparse vectors respectively. \textbf{Bold} indicates performance surpassing the baseline, while \underline{underline} denotes the best performance.}
    \label{tab:NLI_results_uniform_rank}
\end{table*}

\section{mE5$_{\text{BASE}}$}
In Table \ref{tab:SA_results_me5}, we show the results for SA task on mE5$_{\text{BASE}}$. Multilingual E5 text embedding models (mE5) \citep{wang2024multilinguale5textembeddings} are initilizalized from XLM-R models and trained using weakly-supervised contrastive pre-training on billions of text pairs, followed by supervised fine-tuning on a small quantity of high-quality labeled data. Due to the lack of availability of source lang initialization $\phi_L^s$ for mE5$_{\text{BASE}}$, we compare the performance without source lang initialization $\phi_L^s$ for task vectors. We observe an improvement of 2\% using \algo in mE5$_{\text{BASE}}$ model. 

\begin{table*}[t]
    \renewcommand{\arraystretch}{1.1}
    \centering
    \begin{tabular}{c|c|ccccc|c}
    \toprule
        \textbf{Model} & \textbf{Method} & \textbf{mad} & \textbf{bjn} & \textbf{ban} & \textbf{ace} & \textbf{min} & \textbf{Avg.}\\ \midrule
        mE5$_{\text{BASE}}$ & LT-SFT & 59.0 & 82.1 & \underline{73.6} & \underline{68.9} & 77.0 & 72.1 \\ 
        & \algo ($r_l$ = $r_t$ = 100) & \textbf{67.0} & \textbf{83.1} & 70.9 & 64.7 & \underline{\textbf{77.2}} & \textbf{72.6} \\ 
        & \algo ($r_l$ = 90\% var; $r_t$= 90\% var)& \textbf{68.8} & \underline{\textbf{83.5}} & 73.0 & 67.7 & 76.9 & \textbf{74.0} \\
        & \algo ($r_l$ = 100; $r_t$= 90\% var)& \textbf{66.2} & \underline{\textbf{83.5}} & 70.3 & 63.8 & 77.0 & 72.1 \\ 
        & \algo ($r_l$ = 90\% var; $r_t$= 100)& \underline{\textbf{69.3}} & \underline{\textbf{83.5}} & 72.5 & 68.2 & 76.7 & \underline{\textbf{74.1}} \\ \bottomrule
    \end{tabular}
    \caption{Zero-shot cross-lingual transfer evaluation (F1-Score) on SA task (NusaX) using mE5$_{\text{BASE}}$ without $\phi_L^s$ initialization for task. For each model, \textbf{bold} indicates performance surpassing the baseline LT-SFT, while \underline{underline} denotes the best performance.}
    \label{tab:SA_results_me5}
\end{table*}

\end{document}